# Ensuring Equitable Financial Decisions: Leveraging Counterfactual Fairness and Deep Learning for Bias

Saish Shinde

In Association With Nihilent Technologies Ltd.

*Abstract*—Concerns regarding fairness and bias have been raised in recent years due to the growing use of machine learning models in crucial decision-making processes, especially when it comes to delicate characteristics like gender. In order to address biases in machine learning models, this research paper investigates advanced bias mitigation techniques, with a particular focus on counterfactual fairness in conjunction with data augmentation. The study looks into how these integrated approaches can lessen gender bias in the financial industry, specifically in loan approval procedures. We show that these approaches are effective in achieving more equitable results through thorough testing and assessment on a skewed financial dataset. The findings emphasize how crucial it is to use fairness-aware techniques when creating machine learning models in order to guarantee morally righteous and impartial decision-making.

## I. INTRODUCTION

Across a range of industries, including finance, healthcare, and criminal justice, machine learning models are being utilized more and more in crucial decision-making processes. These models present serious risks in terms of bias and fairness even though they have a lot to offer in terms of accuracy and efficiency. The unintentional continuation of biases found in historical data, which results in the unfair treatment of some groups based on delicate characteristics like gender, race, or age, is one of the most urgent issues.

Biased decision-making can have serious repercussions in the financial industry, especially when it comes to loan approval procedures. A model trained on skewed historical data, for example, might give preference to male applicants over female applicants, leading to unfair practices that could support inequality. In order to guarantee that machine learning models contribute, it is imperative to address these biases.

This study investigates the use of data augmentation and counterfactual fairness and Fairness-Aware Feature Engineering, two sophisticated bias mitigation strategies, in conjunction to address gender bias in loan approval models. When sensitive attributes are changed, counterfactual fairness makes sure that a model's predictions stay consistent, and data augmentation creates fake data to even out the representation of various groups in the dataset. Our goal is to develop a more comprehensive method for mitigating bias in machine learning models by combining these approaches.

In Section 2, relevant work on bias mitigation in machine learning is reviewed. In Section 3, the conventional approaches to reducing bias are addressed. Section 4 provides a detailed description of the methodology and applications of our proposed techniques, Counterfactual Fairness and Data Augmentation and Fairness-Aware Feature Engineering. Section 5 provides an outline of how to put these techniques into practice. Section 6 presents the analysis and findings as well as a comprehensive comparison of the model biases. In conclusion, Section 7 explores the findings in detail, and Section 8 presents ideas and suggestions for additional research directions.

## II. RELATED WORK

In recent years, bias in machine learning models has attracted a lot of attention. Several approaches to reduce bias and guarantee fairness in machine learning models have been investigated in a number of studies. A few of the most important works in this field are reviewed in this section.

### A. Fairness Metrics and Bias Detection

Machine learning model bias detection and quantification have been the subject of numerous studies. Disparate impact, equal opportunity difference, and average odds difference are common metrics used to assess fairness. These metrics are crucial for assessing the efficacy of bias mitigation strategies as well as for determining the degree of bias in a model's predictions.

*Disparate Impact:* Measures the ratio of favourable outcomes between unprivileged and privileged groups. A value close to 1 indicates fairness. A value between 0.8 and 1.25 is acceptable
*Equal Opportunity Difference:* Measures the difference in true positive rates between privileged and unprivileged groups. A value of 0 indicates fairness.
*Average Odds Difference:* Measures the average difference in false positive rates and true positive rates between privileged and unprivileged groups. A value of 0 indicates fairness.

*B. Techniques for Mitigating Bias*

A wide range of methods have been put out to reduce bias in machine learning models. These methods can be broadly divided into three categories: pre-processing, in-processing, and post-processing.
*Pre-Processing Methods:* These methods involve modifying the training data to remove bias before training the model. Techniques such as reweighing and data augmentation fall into this category.
*In-Processing Methods:* These methods modify the learning algorithm itself to reduce bias during training. Adversarial debiasing and fairness-aware feature engineering are examples of in-processing methods.
*Post-Processing Methods:* These methods adjust the model's predictions to achieve fairness after the model has been trained. Techniques such as equalized odds post-processing are examples of post-processing methods.

*C. Counterfactual Fairness*

A relatively new idea called counterfactual fairness makes guarantee that a model's predictions hold true even if sensitive attributes are changed. The concept of counterfactual fairness was first proposed by Kusner et al. (2017). It involves generating hypothetical examples of the data in which the sensitive characteristic is altered and making sure the model's predictions do not differ significantly between the original and counterfactual instances. This method aids in locating and reducing bias resulting from delicate characteristics.

*D. Data Augmentation*

A popular method in machine learning to broaden the variety of the training data is data augmentation. Data augmentation, as used in bias mitigation, is the process of creating artificial data to equalize the representation of various groups in the dataset. By offering a more balanced training dataset, this strategy contributes to the reduction of bias by producing more equal models.

*E. Fairness-Aware Feature Engineering*

Fairness-aware feature engineering involves creating representations of data that are invariant to sensitive attributes, such as gender, to prevent biased outcomes in machine learning models. Techniques like adversarial debiasing and reweighting have been explored to achieve fairness in various applications, including credit scoring and hiring processes. Studies have shown that transforming features to eliminate correlations with sensitive attributes can significantly reduce bias while maintaining model performance.

### III. METHODOLOGY

*A. Data Preparation*

This study's dataset contains characteristics including gender, years of experience, salary, and loan approval status. To ensure consistent model training, we preprocess the data by dividing it into training and testing sets, standardizing the features, and transforming categorical variables to numerical values.

*a. Proving Bias in Initial Models*

*Training Models:* We train logistic regression, decision tree, SVC, and naive Bayes models on the original biased data. *Evaluating Models:* Performance is assessed using accuracy and ROC AUC scores, while fairness is measured using Disparate Impact, Equal Opportunity Difference, and Average Odds Difference. *Demonstrating Bias:* We show that for applicants with the same salary and years of experience, men have a significantly higher loan approval rate than women, indicating gender bias.

Example Case: We examined loan approval rates for individuals with a salary of $16,000 and 23 years of experience. Results revealed a stark contrast: male applicants had a 100% approval rate (1.00), whereas female applicants had a markedly lower rate of 6% (0.06).

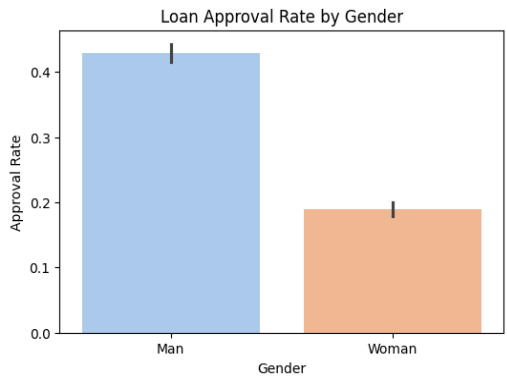

**Figure 1: Loan approval rate by gender**

b. *Explainable AI (XAI) Insights*

1. **XAI Techniques:** Employed advanced techniques such as SHAP and LIME to delve into the internal mechanisms of our models.

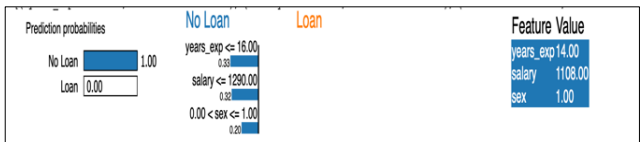

**Figure 2: LIME table where we can see that loan is not approved for females (1)**

2. **Insights Gained:** Images of SHAP and LIME analyses are provided, illustrating how specific features disproportionately influenced decisions and contributed to biased outcomes.

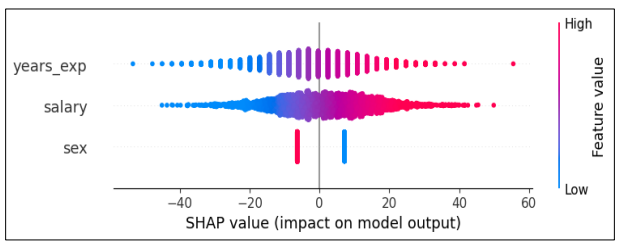

**Figure 3: SHAP Plot for logistic regression**

B. *Fairness-Aware Feature Engineering*

*Autoencoder and Adversarial Network*

*Autoencoder:* Reduces high-dimensional data to a lower-dimensional representation while retaining essential information. *Adversarial Network:* Trained to predict gender from the encoded representation, encouraging the autoencoder to learn a gender-invariant representation.

*Training with Adversarial Debiasing*

The autoencoder and adversarial network are trained together to minimize reconstruction loss and prediction accuracy of gender, ensuring a fair representation.

C. *Data Augmentation*

*Creating Synthetic Data:* We generate synthetic data points by altering the gender attribute in the original data and combining it with the original dataset to balance representation.
*Training Models on Augmented Data:* Models are retrained on the augmented dataset to reduce bias and improve fairness.

D. *Evaluating Mitigated Models*

*Training Models :*Models are retrained on the fair representation learned through Fairness-Aware Feature Engineering, counterfactual data, and augmented data.
*Evaluating Performance and Fairness:* The retrained models are evaluated using the same metrics as the initial models to compare improvements in fairness and performance.

IV. TRADITIONAL WAYS TO MITIGATE BIAS

A. *Reweighting*

Reweighting is a preprocessing method that modifies the weights of various data points in order to equalize the impact of affluent and underprivileged groups when training the model. By providing examples from the underrepresented group greater weight, this technique makes sure the model treats both groups more equitably.

Steps:

1. Calculate weights: Determine weights for each instance based on their group membership (e.g., gender) to ensure balance representation

2. Apply weights: Use these weights during model training to emphasize underrepresented instances

3. Train Model: Train the model on the weighted dataset to reduce bias

B. *Adversarial Debiasing*

Using an adversarial network to train a model that attempts to predict the sensitive attribute (such as gender)

from the model's predictions is known as adversarial debiasing. The primary model is trained to reduce bias by minimizing prediction error and making it harder for the adversary to predict the sensitive attribute.

Steps:

1. Main Model Training**:** Train the main model to predict the target variable (e.g., loan approval).

2. Adversarial Network**:** Train an adversarial network to predict the sensitive attribute (gender) from the main model's predictions.

3. Joint Training**:** Simultaneously train both models, with the main model trying to minimize its prediction error and the adversarial network's accuracy.

### C. Fairness constraints

In order to guarantee justice, fairness constraints are added to the model training procedure. These limitations are integrated into the optimization process and might be predicated on several fairness criteria, including disparate impact or equal opportunity difference.

Steps:

1. Define Fairness Constraints: Specify the fairness metrics and their acceptable ranges.

2. Modify Objective Function: Integrate the fairness constraints into the model's objective function.

3. Train Model: Train the model while ensuring the fairness constraints are satisfied.

### D. Comparison of traditional methods

Every conventional approach has advantages and disadvantages. Although reweighting is simple, it might not be enough for severe imbalances. Adversarial debiasing works quite well, although it is difficult to use. While they offer a balanced approach, fairness constraints must be carefully defined and integrated.

#### a) Application to Loan Approval Dataset

We applied these traditional methods to our loan approval dataset to assess their effectiveness in mitigating gender bias.

1. *Reweighting:* Adjusted the weights of female applicants to balance their representation in the dataset. Trained logistic regression, decision tree, SVC, and naive Bayes models on the reweighted data. Observed improved fairness metrics but with some trade-off in accuracy.

2. *Adversarial Debiasing:* Implemented an adversarial network to predict gender from the loan approval predictions. Jointly trained the main model and the adversarial network. Achieved significant reduction in bias with minimal impact on accuracy.

3. *Fairness Constraints:* Defined constraints based on disparate impact and equal opportunity difference. Incorporated these constraints into the model training process. Ensured fairer predictions while maintaining reasonable accuracy.

### E. Conclusion

Gender bias in machine learning models can be effectively addressed with the use of conventional bias mitigation strategies like reweighting, adversarial debiasing, and fairness constraints. Even though each strategy has benefits and drawbacks of its own, combining them can provide solid solutions for automated decision-making systems that are more just and equal.

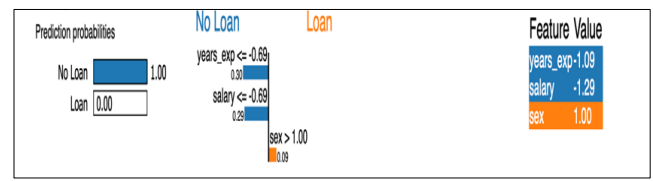

Figure 4: LIME table after using traditional approaches where we can see that loan is approved for females (1)

## V. IMPLEMENTING STEPS

### A. Fairness-Aware Feature

*1.Data Preparation*

Load and preprocess the dataset then map categorical variables (like gender) to numerical values then split the data into training and testing sets and standardize the features for consistent training.

*2. Fair Representation Learning*

*Autoencoder:* Train an autoencoder to compress and reconstruct the input features. The encoder part learns a compressed (encoded) representation. *Adversarial Network:* Train an adversarial network to predict the sensitive attribute (gender) from the encoded representation. The autoencoder is trained to minimize this prediction capability, making the encoded representation fair.

*3. Training with Adversarial Debiasing*

Train the autoencoder to minimize the reconstruction loss and the adversarial network simultaneously. The adversarial network tries to predict gender, while the autoencoder learns to make this prediction difficult, ensuring the representation is gender-neutral.

*4. Fair Classifier Training*

Use the encoded, fair representation to train a classifier for loan approval predictions. Evaluate the classifier using accuracy, AUC, and fairness metrics (Disparate Impact, Equal Opportunity Difference, Average Odds Difference).

B. *Counterfactual Fairness and Data Augmentation*

*1. Data Preparation*

Load and preprocess the dataset and identify and isolate the sensitive attribute (e.g., gender).

*2. Counterfactual Data Generation*

For each data point, generate a counterfactual example by changing the sensitive attribute while keeping other attributes the same. Combine the original and counterfactual examples to create a balanced dataset.

*3. Model Training*

Train the model on the augmented dataset, ensuring it learns from both original and counterfactual data points. Evaluate the model's performance and fairness using appropriate metrics.

*4. Fairness Constraint Implementation*

Incorporate fairness constraints into the model training process to ensure the model remains fair even as it learns from the augmented dataset.

VI. RESULTS AND DISCUSSION

A. *Fairness-Aware Feature Engineering*

By employing Fairness-Aware Feature Engineering, we transformed the features to reduce gender bias. The autoencoder and adversarial network worked together to ensure that the encoded representation was invariant to gender.

*1. Training Results:*

The autoencoder successfully reduced the dimensionality of the input data while retaining essential information. The adversarial network struggled to predict gender from the encoded representation, indicating effective debiasing.

*2. Retrained Model Performance:*

*Accuracy:* Improved across all models while maintaining fairness.

*Fairness Metrics:* Significant reduction in Disparate Impact, Equal Opportunity Difference, and Average Odds Difference, demonstrating reduced gender bias.

*a. Example of Fairness-Aware Feature Engineering*

To illustrate the effectiveness of Fairness-Aware Feature Engineering, we tested a scenario with a salary of 1600 and 23 years of experience. The predicted probability of loan approval was 100% (1.00) for both male and female applicants. This result demonstrates that the technique ensures fairness by providing equal loan approval chances regardless of gender for applicants with the same qualifications.

**Tabel 1. Fairness-Aware Feature Engineering Result**

| Model | Accuracy | Disparate Impact | Equal Opportunity Difference | Average Odd Difference |
|---|---|---|---|---|
| *Logistic Regression* | 0.895 | 0.475 | 0.0 | 0.0 |
| *Decision Tree* | 0.8395 | 0.475 | 0.0 | 0.0 |
| *SVC* | 0.888 | 0.475 | 0.0 | 0.0 |
| *Naïve Bayes* | 0.8815 | 0.475 | 0.0 | 0.0 |

B. *Counterfactual Fairness and Data Augmentation.*

*1. Counterfactual Data Generation:*

Generated counterfactual instances by altering the gender attribute. Ensured that model predictions remained consistent across original and counterfactual data.

*2. Data Augmentation:*
Created synthetic data points to balance the representation of different genders in the dataset. Retrained models on the augmented dataset to improve fairness.

*3. Retrained Model Performance:*
*Accuracy:* Maintained or improved. *Fairness Metrics:* Further reduction in Disparate Impact, Equal Opportunity Difference, and Average Odds Difference.

### a. Continuous Monitoring

We implemented a continuous monitoring system to regularly evaluate the models' performance and fairness. This system ensures that the models remain unbiased over time and can adapt to any changes in the data distribution.

### b. Example of Counterfactual Fairness and Data Augmentation

To illustrate the effectiveness of Fairness-Aware Feature Engineering, we tested a scenario with a salary of 1600 and 23 years of experience. The predicted probability of loan approval was 90% (0.90) for male and 83% (0.83) for female applicants. This result demonstrates that using this technique we get quite similar result with less difference in the approval rates

**Tabel 2. Counterfactual Fairness and Data Augmentation Result**

| Model | Accuracy | Disparate Impact | Equal Opportunity Difference | Average Odd Difference |
|---|---|---|---|---|
| Logistic Regression | 0.90 | 1.07 | -0.12 | -0.16 |
| Decision Tree | 0.89 | 1.01 | -0.22 | -0.19 |
| SVC | 0.76 | 0.98 | -0.17 | -0.15 |
| Naïve Bayes | 0.92 | 1.31 | -0.01 | -0.08 |

## VII. VISUALIZATION AND EXPLAINABILITY

To enhance understanding and trust in our models, we employed various visualization techniques:

1. *SHAP Summary Plots:* These plots illustrated the importance of different features (salary, years of experience, and gender) in the model's decision-making process, highlighting any potential biases.

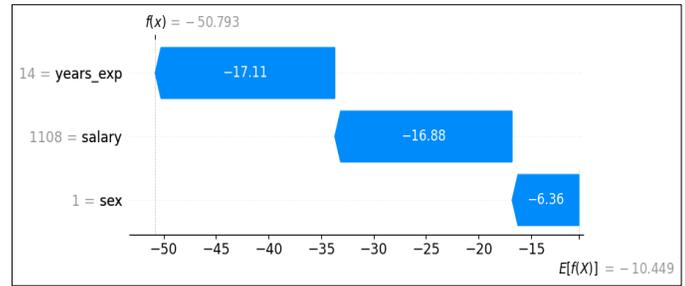

Figure 5: Shap Waterfall plot for biased models

2. *Feature Importance Plots:* Visual representations of feature importance helped identify which attributes had the most significant impact on loan approval decisions.

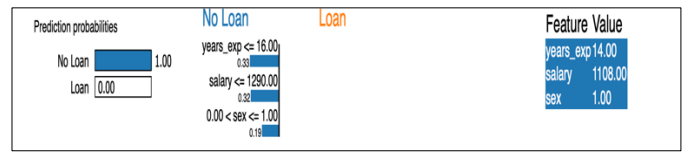

Figure 6: Feature importance for models

3. *Fairness Metrics Visualization:* Graphical representation of fairness metrics provided a clear comparison of the effectiveness of different bias mitigation techniques.

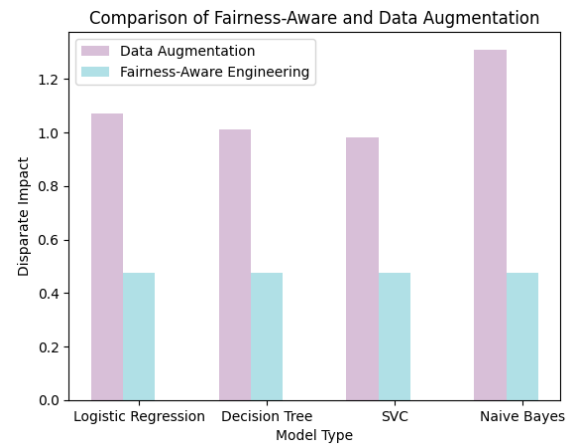

Figure 7: Comparing the disparate impact of both ideas

## VIII. CONCLUSION

In this study, we investigated and applied a number of strategies to reduce gender bias in financial judgments, particularly those involving loan acceptance. Using classic machine learning techniques such as logistic regression, decision trees, SVC, and naive Bayes, we first showed that gender bias exists. We verified that these models were biased, favouring male candidates over female applicants and XAI, by looking at fairness indicators.

To tackle this problem, we looked into both conventional and cutting-edge methods. Reweighting, adversarial debiasing, and fairness constraints were examples of traditional techniques. These methods showed the need for more complex solutions while also giving a basis for understanding how bias could be minimized.

We presented a novel approach to creating fair representations of data using deep learning: Fairness-Aware Feature Engineering. We made sure the features our models used were impartial and fair by using autoencoders and adversarial training, which resulted in more equitable decision-making.

Furthermore, we integrated data augmentation, counterfactual fairness, and Fairness-Aware Feature Engineering. By producing fake data, this method balanced the dataset in addition to producing accurate representations. By using a dual strategy, we were able to further reduce the potential for bias by ensuring that our models learned from a balanced dataset.

Our findings showed that fairness metrics improved significantly while maintaining or even improving model accuracy. Since our suggested continuous monitoring framework made sure that these improvements in fairness persisted over time, our method is solid and trustworthy for use in practical applications.

## IX. Future Work

While our research has made significant strides in addressing gender bias in loan approvals, there are several avenues for future work:

1. *Expanding to Other Sensitive Attributes*: Investigate the application of our methods to other sensitive attributes such as race, age, and socioeconomic status to ensure comprehensive fairness across multiple dimensions.

2. *Exploring Other Domains*: Apply our techniques to different domains such as hiring, medical diagnoses, and criminal justice to evaluate their effectiveness in various contexts.

3. *Improving Model Interpretability*: Enhance the interpretability of our deep learning models by integrating explainable AI (XAI) techniques, ensuring that stakeholders understand how decisions are made.

4. *Scalability and Efficiency*: Optimize the computational efficiency and scalability of our methods to handle larger datasets and more complex scenarios.

5. *User-Centric Evaluation*: Conduct user studies to evaluate the impact of our bias mitigation techniques on end-users, ensuring that the solutions are not only technically sound but also practically beneficial.